\documentclass{article}

\usepackage{microtype}
\usepackage{graphicx}
\usepackage{subfigure}
\usepackage{booktabs} 
\usepackage{makecell}
\usepackage{hyperref}

\newcommand{\ty}[1]{{\color{violet} #1}}

\usepackage[accepted]{icml2024/icml2024}

\usepackage{amsmath}
\usepackage{amssymb}
\usepackage{mathtools}
\usepackage{amsthm}
\usepackage{multirow}
\usepackage{enumitem}

\usepackage[capitalize,noabbrev]{cleveref}

\theoremstyle{plain}

\theoremstyle{definition}

\theoremstyle{remark}

\usepackage[textsize=tiny]{todonotes}

\icmltitlerunning{Sub-token ViT Embedding via Stochastic Resonance Transformers}

\begin{document}

\twocolumn[
\icmltitle{Sub-token ViT Embedding via Stochastic Resonance Transformers}



\icmlsetsymbol{equal}{*}

\begin{icmlauthorlist}
\icmlauthor{Dong Lao}{UCLA}
\icmlauthor{Yangchao Wu}{UCLA}
\icmlauthor{Tian Yu Liu}{UCLA}
\icmlauthor{Alex Wong}{Yale}
\icmlauthor{Stefano Soatto}{UCLA}
\end{icmlauthorlist}

\icmlaffiliation{UCLA}{UCLA Vision Lab}

\icmlaffiliation{Yale}{Yale Vision Lab}

\icmlcorrespondingauthor{Dong Lao}{lao@cs.ucla.edu}


\vskip 0.3in
]



\printAffiliationsAndNotice{}  

\begin{abstract}
Vision Transformer (ViT) architectures represent images as collections of high-dimensional vectorized tokens, each corresponding to a rectangular non-overlapping patch. This representation trades spatial granularity for embedding dimensionality, and results in semantically rich but spatially coarsely quantized feature maps. In order to retrieve spatial details beneficial to fine-grained inference tasks we propose a training-free method inspired by  ``stochastic resonance.'' Specifically, we perform \emph{sub-token spatial transformations} to the input data, and aggregate the resulting ViT features after applying the 
\emph{inverse transformation}. The resulting  ``Stochastic Resonance Transformer" (SRT) retains the rich semantic information of the original representation, but grounds it on a finer-scale spatial domain, partly mitigating the coarse effect of spatial tokenization.  
SRT is applicable across any layer of any ViT architecture, consistently boosting performance on several tasks including segmentation, classification, depth estimation, and others by up to 14.9\% without the need for any fine-tuning. Code: \url{https://github.com/donglao/srt}.

\comment{We describe a simple method to use Vision Transformers (ViTs) to represent images at a level of granularity below that afforded by the tokenization scale. We use so-called Stochastic Resonance, an averaging technique common in audio processing to increase the quantizer resolution by averaging artificially perturbed data. In a ViT, we introduce controlled perturbations and aggregate the resulting embeddings to produce  sub-token level representations without altering network architecture or pre-trained weights. This allows using our approach on any pre-trained ViT model. We test our approach in transductive learning, adapting pre-trained models ({\em e.g.} DINO) to new image domains without ground truth annotations and achieve significant improvements in various zero-shot vision tasks. Optionally, we  distill  fine-grained features back into the original ViT resolution, which still improves performance without increasing inference cost and latency.}

\end{abstract}

\section{Introduction}
The Transformer architecture \cite{vaswani2017attention}, originally designed for modeling language which is naturally quantized into discrete objects (sub-word ``tokens''), is a poor fit for vision tasks due to the lack of a natural scale for spatial discretization: The same object can disappear within a pixel or fill the entire image plane depending on its distance from the camera. In theory, one could create tokens for patches of all sizes and positions, but at significant computational expense due to the complexity of transformers, which is quadratic in the number of tokens. Despite the counter-intuitive nature of spatial quantization, Vision Transformers (ViTs) \cite{dosovitskiy2020image} achieve state-of-the-art performance in many vision tasks. So we focus on developing methods to harness pre-trained ViTs and overcome their limitations in representing spatial details at fine granularity due to  the fixed spatial quantization of tokens.

The standard remedy for quantization artifacts is anti-aliasing. In one-dimensional digitized signals such as audio, anti-aliasing refers to weighted averaging of nearby samples in the discrete topology, or equivalently averaging versions of the signal translated by {\em integer multiples of the original sampling interval}. For images, in addition to sampling the translation group, one also has to sample the scale group, so as to capture the varying size of the projection of objects onto the image plane. Various network architectures comprise spatial average pooling, which is translational anti-aliasing, whereas the notion of domain-size pooling and anti-aliasing has been championed by \citep{dong2015domain}. Anti-aliasing is typically performed by convolving the discrete signal with a generic (not data-dependent) kernel. The optimal kernel is unbounded, so any finite implementation is necessarily lossy and cannot ``recreate information'' lost in the sampling process. Similarly, super-resolution algorithms hallucinate missing details either using generic priors or data other than the signal in question \cite{buades2005non}. 

``Stochastic Resonance'' \cite{benzi1981mechanism}  is a qualitatively different process whereby the limitations imposed by a fixed quantization threshold can be overcome simply by shifting the signal by sub-threshold additive perturbations. This results in sampling beyond the Nyquist limit otherwise imposed by the quantizer. We extend this process, originally employed in cochlear implants, to translation {\em not} of the value of the signal (additive perturbations) but its domain (translation). The same process can also be applied to domain size (scale). 
Stochastic resonance can be thought of as a form of data augmentation or adaptive quantization. We further simplify it by choosing deterministic, rather than randomly sampled, perturbations.  The perturbed token embeddings are aggregated statistically to first-order (mean or median) 
to yield a sub-token embedding. 
First-order statistics can be used for visual tasks, for instance unsupervised object segmentation, and second-order statistics as a weight for adaptive regularization. 

This simple approach is well suited to pre-trained transformers since it only requires acting on inputs and outputs without modifying (or even knowing) the weights or the forward pass of the model. 
\begin{figure*}[t]
\centering
\includegraphics[width=0.8\textwidth]{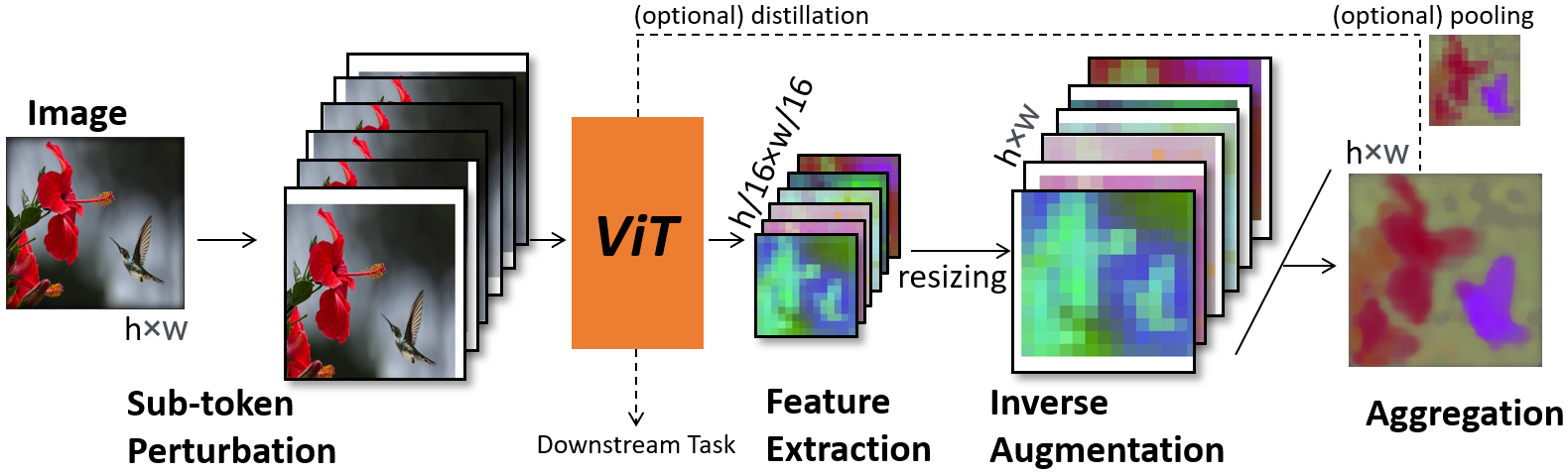}

    \caption{\sl\small {\bf Schematic for SRT. } 
    SRT applies controlled perturbations to input images, extracting features through Vision Transformers (ViTs). These features are then upsampled to higher resolution and aligned using the inverse of the applied perturbations. Statistical aggregation, including mean and median, along the perturbation dimension, produces fine-grained feature representations. These features find utility in visualization and can also be seamlessly integrated back into the network for enhanced performance in downstream tasks.
    }
    \label{fig:schematic}
\end{figure*}
We call the resulting method ``Stochastic Resonance Transformer'' although we do not modify the transformer nor do we use artificial noise, to reflect closer proximity of our method to Stochastic Resonance than to traditional super-resolution or anti-aliasing methods. The simplicity of the method allows us to leverage ViTs, pre-trained on large datasets, such as CLIP and DINO, to improve their handling of spatial quantization. This may help attenuate some of the biases of these datasets, for instance, the object-centric nature of DINO, which biases the representation towards centered objects that occupy a large portion of the visual field. Stochastic Resonance  can be used as a form of sub-token ensembling, to enhanced feature maps in ViTs and reveal some of the local fine-grained underlying structure.
SRT can be applied to any ViT layer, on any task, without altering network architecture or pre-trained network weights. 
We can use SRT to visualize fine-grained features, or optionally map them back to the original ViT feature scale by pooling to be used for inference, where we notice performance increases on a wide range of vision tasks. Additionally, fine-tuning pre-trained ViTs by distillation from ensembled features maintains their original inference time and cost. To the best of our knowledge, SRT is the first approach to recoup spatial granularity from embedding dimension in ViT feature maps. Unlike conventional ensemble methods that augment model inputs and combine outputs, SRT operates at the feature level, allowing seamless integration into any ViT pipelines,  including for tasks that demand intermediate features or attention maps. Our contributions are summarized as follows:

\setlist{nosep}
\begin{itemize}[leftmargin=*]
\item We introduce a novel technique, namely the Stochastic Resonance Transformer (SRT), that computes fine-grained ViT embeddings at test time without additional training or modifications to the ViT's forward pass.
\item SRT can be seamlessly integrated into any task that utilizes ViT as a feature extractor, serving as a test-time augmentation and ensemble method.
\item We provide an efficient implementation SRT, including parallelization and recursive aggregation, which reduces computational and memory requirements.
\item We showcase the effectiveness of SRT by consistent improvement on a range of diverse vision tasks. Notably, it demonstrates significant enhancements on dense prediction tasks, of up to 14.9\% on depth prediction.
\item SRT also yields a versatile visualization tool that can be applied to any layer of any pre-trained ViT model, offering valuable insights into ViT model characteristics.
\end{itemize}

\comment{
In Sect.~\ref{sec:implementation}, We first presentation the conceptual framework and pipeline of SRT, then we formalize SRT in Sect.~\ref{sec:theory}, and present an efficient implementation in Secton~\ref{sec:efficient}. We demonstrate the effectiveness of SRT on 7 different vision tasks, 5 in Sect.~\ref{sec:experiments} and 2 in the Appendix. Finally, we review literature related to SRT in Sect.~\ref{sec:related}, and provide a discussion on limitations in Sect.~\ref{sec:conclusion}.  
}

\section{Stochastic Resonance Transformer}
\subsection{Method}\label{sec:implementation}

Given an image $x$ with $N\times M$ resolution, a Vision Transformer (ViT) divides it into tokens, where each token represents a $n\times m$ rectangular patch. While tokens can technically overlap, practical ViT models often use non-overlapping tokens for efficiency due to the quadratic complexity of transformers with respect to the number of tokens. Consequently, in a certain layer of ViT, this approach yields a feature map with dimensions $\frac{N}{n}\times \frac{M}{m}\times C$, where $C$ is the size of the feature vector determined by architecture, downsampled from the original image and subsequently losing sub-token spatial information.

Given a trained ViT model, we aim to obtain features in a higher resolution that preserves the spatial information on a pixel level, ideally matching with the original image input. Fig.~\ref{fig:schematic} illustrates our proposed pipeline of SRT. To enhance the features, we introduce sub-token perturbation to the input, i.e. transforming the coordinates of the input and resampling onto a new image plane, and extract embeddings from the resulting perturbed image. Note that, any group transformation (translation, rotation, flipping, etc.)
and a combination of them ({\em e.g.} \cite{wu2023augundo}) can be chosen as the perturbation, provided that its inverse transformation is available. However, our specific interest lies in introducing perturbation through translation. This preference arises from several factors: 1) translation preserves the object's scale, unlike zooming; 2) translation can be applied at the pixel level grid, eliminating interpolation artifacts, as opposed to rotation; 3) translation allows an efficient implementation of SRT (Sect.~\ref{sec:efficient}).

We then upsample the resulting low-resolution embeddings back to the original image resolution $N\times M$ and apply an inverse of the perturbation to the spatial coordinates of the embeddings, and through an inverse warp, align it with the original input image.
By repeating this process on different sub-token perturbations for $t$ times, we generate a collection of embeddings, denoted by  $N\times M \times C \times t$, that are spatially aligned to the input frame of  reference. We can then compute statistics, {\em e.g.} mean or median, along the $t$ dimension. Consequently, we obtain a feature field $N\times M \times C$, with the same spatial resolution as the original image. As showcased in Fig.~\ref{fig:vits}, the embeddings are enhanced to sub-token resolution. 
This process is similar to Stochastic Resonance, where introducing white noise to the input signal enhances a signal beyond the native resolution. These embeddings offer promising downstream applications, as in Sect.~\ref{sec:experiments}.

For any task that utilizes ViT as a feature extractor, we can take an additional step by applying average pooling to again tokenize this high-resolution feature, to map it to $\frac{N}{n}\times \frac{M}{m}\times C$. It's important to note that this feature differs from the one obtained from one single forward pass of ViT, as it is an aggregate 
of multiple perturbed inputs. This process can be viewed as test-time augmentation and ensemble. Since this feature is compatible with the original ViT architecture, it can be seamlessly integrated back into the layer from which we perturbed the features, and is applicable to any model at any layer, regardless of pre-training, without requiring additional learned modules or altering the forward pass. Such a pipeline improves performance on diverse computer vision tasks, as validated by Sect.~\ref{sec:experiments}. Next, we formalize the aforementioned pipeline.

\subsection{Formalization}~\label{sec:theory}
$x \in {\mathbb R}^{N\times M\times K}$ is a $K$-channel signal ({\em e.g.,} $K = 3$ for a color image.) Let $\pi: {\mathbb R}^{N\times M} \rightarrow {\mathbb R}^{n\times m}; x \mapsto x$ a projection (subsampling, $n \ll N, m \ll M$), with the corresponding inverse (interpolation) map $\pi^{-1}: {\mathbb R}^{n\times m} \rightarrow {\mathbb R}^{N\times M}; x \mapsto x$ be piecewise constant. This is a trivial form of subsampling and interpolation with a constant kernel.

Now, let $\phi: {\mathbb R}^{NMK} \rightarrow {\mathbb R}^{nmC}$ a trained model with $C$ channels of feature maps, typically $C \gg K$. Finally, let $T: {\mathbb R}^{N\times M} \rightarrow {\mathbb R}^{N\times M}; x \mapsto Tx$ a compact and invertible transformation, for instance, edge-padded shift by a number of pixels smaller than $(N-n)/n\times (M-m)/m$. We consider uniform random padded shifts (translation) and consider the following measurement process:
\begin{equation}
    y_t = \phi(T_t x)
\end{equation}
for all random transformations $T_t$. We wish to enhance the output of $\phi$ from $n\times m$ to $N\times M$. We call this process {\em immersion} since  each point $x$ maps to $z = \phi(x)$ but $z \neq T^{-1} \phi(Tx)$. In other words, $x$ is mapped injectively but not bijectively, since there are as many (vector)-values as the sampled value of $T$. We do so iteratively by averaging (or by a linear transformation $K_t$) with respect to the innovation process:
\begin{equation}
    \epsilon_t = \underbrace{\pi\left(T_t^{-1} \pi^{-1} y_t\right)}_{\hat y_t} - K_t \phi(x)
\end{equation}
now the fine-grained features which we call $\hat x_t$ are obtained by an observer architecture, which implements a closed-loop dynamical system of the form:
\begin{equation}
    \begin{cases}
        \hat x_{t+1} = \hat x_t + T_t^{-1} \pi^{-1} y_t \quad \hat x_0 = 0; \\
        y_t = \phi(T_t x)
    \end{cases}
\end{equation}
This is just a moving average in higher resolution, whereby the variance of $\hat x$ will decrease to a steady state (by Central Limit Theorem), following the practice of stochastic resonance. It is a mixture of upsampling/interpolation and inference-time data augmentation, or ensembling. \footnote{A detailed formalization of the image quantization artifacts is deferred to Appendix~\ref{sec:formalization_appendix}.}

\comment{
Now we also want to infer $K_t$ transductively so that the innovation is minimized. The supervisory signal, manifest at test time, is in the form of a task-specific functional ({\em e.g.}, a segmentation functional $h$) that takes as input both the original data $x$ and the super-resolved feature $\hat x$ that represents past data. Therefore, the processing of the present data is done through optimization that involves both past data reflected in the training set used to instantiate $\phi$ that produces $\hat x$, as well as current data $x$. We denote the functional as $r(x, \hat x)$. This indicates some kind of residual or loss.  The most obvious loss or residual is:
\begin{equation}
    r_t(x, \hat x) = \hat x_t - \pi^{-1}( K \phi(x)).
\end{equation}
Optimizing this with respect to $K$ amounts to transductive data augmentation of transductive contrastive learning, which yields $\hat K(x)$:
\begin{equation}
    \hat K(x) = \arg\min_K \sum_{t = 1}^{\tau}  r_t(\hat x_t, x)
\end{equation}
Averaging the resulting $K(x)$ for a fine-tuning dataset yields contrastive regularization of the pre-trained embedding $\phi$. Alternatively, one could extend $K$ to encompass all parameters of $\phi$ instead of just the last linear layer as suggested above.

\subsection{Extension}

The resulting regularized embedding can be fed to a transformer just as the original features were. It can be used for segmentation where the original feature map can guide a trained transformer-based segmentation, and the super-resolved feature map can guide the traditional energy-based segmentation of the original image (guided segmentation).

\subsection*{Caveat emptor}
Q: How is this not just data agumentation that starts after a few epochs of (pre-)training?

A: this is a transducive form of regularization taht can be used to perform task-specific and sample-specific inference. But this needs to be done in a sound manner, and yield compelling results.

\subsection{Implementation}

\begin{itemize}
    \item Feature level alignment:
    \item Sample from uniform distribution, use random translations 

    \item Fast implementation
    \item parallel computing is possible through batching
    \item ideally the fine-grained feature should be directly used for downstream task, however due to architecture constraints, we may need to do pooling to downsample.
    \item However, this may speed up inference, as when combined with average pooling one can by-pass the memory-demanding feature resizing step.
\end{itemize}
}
\def\figd{figures/vis_vit}
\def\fWidD{0.11\textwidth}

\begin{figure*}[t!]
\hspace*{-0.1in}
\centering
{\footnotesize 
\begin{tabular}{c@{\hspace{0.2in}}c@{\hspace{0.05in}}c@{\hspace{0.05in}}c@{\hspace{0.05in}}c@{\hspace{0.05in}}c@{\hspace{0.05in}}c}
\smash{{\raisebox{-10mm}{Input}}}& &\thead{ \scriptsize Visual-language\\\scriptsize CLIP} & \thead{\scriptsize Classification\\\scriptsize Supervised} & \thead{\scriptsize Contrastive\\\scriptsize DINO} & \thead{\scriptsize Segmentation\\\scriptsize SAM} & \thead{\scriptsize Reconstruction\\\scriptsize MAE}
\\
\smash{\raisebox{-10mm}{\includegraphics[width=\fWidD]{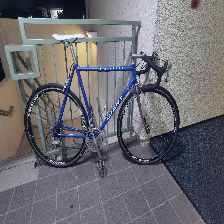}}}&\rotatebox{90}{\quad ViT Feature}&\includegraphics[width=\fWidD]{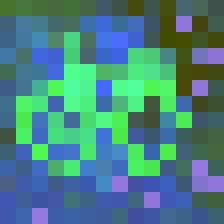} & \includegraphics[width=\fWidD]{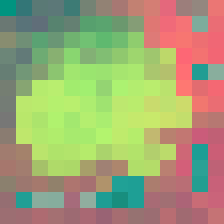} & \includegraphics[width=\fWidD]{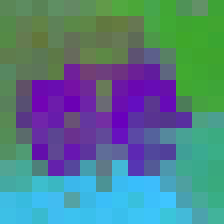} & \includegraphics[width=\fWidD]{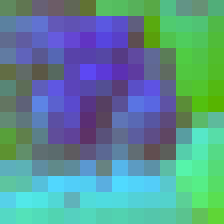} & \includegraphics[width=\fWidD]{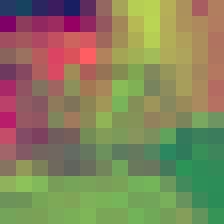} \\&\rotatebox{90}{\quad\,\, + SRT}&
\includegraphics[width=\fWidD]{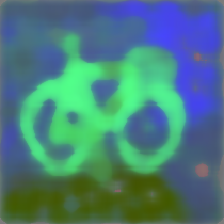} & \includegraphics[width=\fWidD]{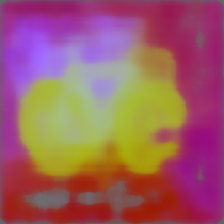} & \includegraphics[width=\fWidD]{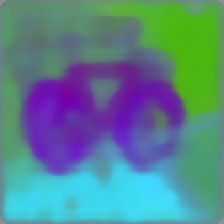} & \includegraphics[width=\fWidD]{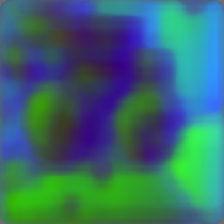} & \includegraphics[width=\fWidD]{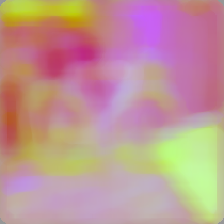} \\
\smash{\raisebox{-10mm}{\includegraphics[width=\fWidD]{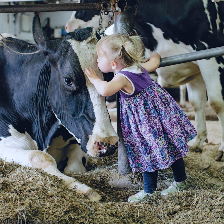}}}&\rotatebox{90}{\quad ViT Feature}&\includegraphics[width=\fWidD]{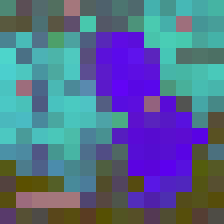} & \includegraphics[width=\fWidD]{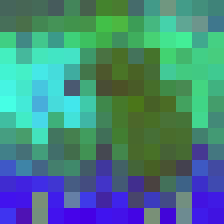} & \includegraphics[width=\fWidD]{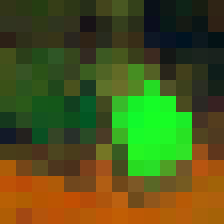} & \includegraphics[width=\fWidD]{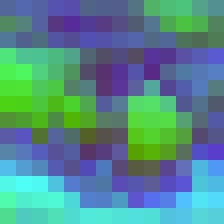} & \includegraphics[width=\fWidD]{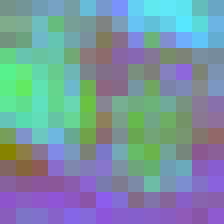} \\&\rotatebox{90}{\quad\,\, + SRT}&\includegraphics[width=\fWidD]{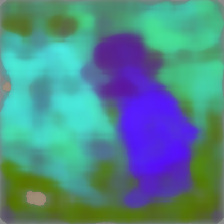} & \includegraphics[width=\fWidD]{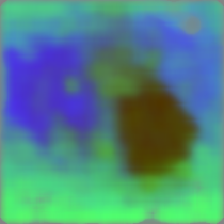} & \includegraphics[width=\fWidD]{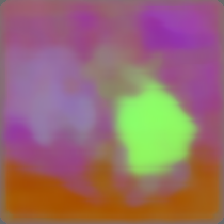} & \includegraphics[width=\fWidD]{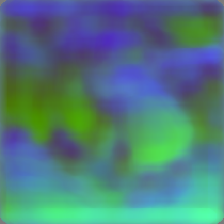} & \includegraphics[width=\fWidD]{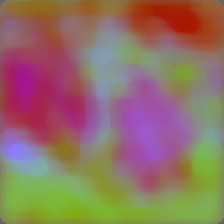} \\
\end{tabular}
}

\caption{\sl\small {\bf High-resolution ViT features computed by stochastic resonance.} Stochastic Resonance enables enhancing tokenized ViT features during inference without the need for additional training or modifying ViT forward pass. Here we present enhanced features from different pre-trained ViT models, visualized via Principal Component Analysis: CLIP \citep{radford2021learning} captures major image components. Interestingly, although Supervised \citep{dosovitskiy2020image} and DINO \citep{caron2021emerging} are trained by different pipelines and training loss, they prioritize similar regions. This may be due to they are trained on the same dataset and thus capture similar inductive bias. In contrast, SAM \citep{kirillov2023segment} and MAE \citep{he2022masked} capture local features over high-level semantics.}
\label{fig:vits}
\end{figure*}

\subsection{Efficient Implementation}\label{sec:efficient}
In theory, there is no limitation on the types of sub-token transformations that can be employed. We opt for a straightforward approach by applying translations (with padding) and this practice demonstrates effective results. We sample translations at the pixel level, avoiding the need for sub-pixel interpolation, which could introduce unwanted artifacts.

For a ViT utilizing token sizes of $m \times n$, we impose a constraint on the maximum magnitude of translation, limiting it to $\frac{m}{2}\times \frac{n}{2}$. This constraint allows the model to explore all possible token selections within the image. It is worth noting that excessive translation can be counterproductive when applied to downstream vision tasks, as it can result in information loss at the image boundaries. A detailed discussion can be found in Sect. \ref{sec:davis}, where we study the relation between perturbation level and model performance.

While naive implementation can lead to significant computational drawbacks, running inference on each augmented image can be trivially parallelized. Greater implementation speed-ups can also be achieved by bypassing the upsampling step (which is computationally expensive), since the aggregated result can be deterministically computed from the original feature maps of each augmented image when average pooling is used. With ViT-16/S architecture, on DAVIS-2017 \citep{Pont-Tuset_arXiv_2017} our implementation of SRT runs at 1.0 seconds per image on a Nvidia 3090 GPU using a perturbation level of 3 pixels. To further speed up, one may optionally fine-tune the ViT model by distilling utilizing SRT, so that the inference time and cost remain, as demonstrated in Sect.~\ref{sec:davis}. We include the demo code in the supplementary material and will make it publicly available.

\section{Experiments}\label{sec:experiments}


\subsection{Visualization of SRT Features}
SRT demonstrates significant promise in visualizing features of ViT models. It achieves this without necessitating modifications to the ViT's forward pass. In Fig.~\ref{fig:vits}, we present visualizations of the final layer features from five popular ViT models, all employing the ViT-B/16 architecture. Notably, all visualizations are computed by a standard consumer laptop. We employ SRT with a turbulence level of 7 pixels to traverse non-overlapping augmented tokens extensively. The resultant high-dimensional features then go through Principal Component Analysis (PCA), with the top three components mapped to RGB channels to facilitate effective visualization. Despite sharing the same architecture, the five models exhibit distinct characteristics owing to variations in their pre-training supervision. For instance, CLIP \citep{radford2021learning} is trained through contrastive visual-language pre-training and captures major image components in the displayed examples. The Supervised model \citep{dosovitskiy2020image} is trained for ImageNet classification, while DINO \citep{caron2021emerging} undergoes contrastive learning. Interestingly, despite their diverse training regimes, both models prioritize similar image regions, potentially due to their shared dataset and resulting common inductive bias. In contrast, SAM \citep{kirillov2023segment} is trained on massive segmentation masks without semantic labels or object-centric priors, and MAE \citep{he2022masked} is trained through inpainting of randomly masked image regions. Both methods emphasize local image features over high-level semantics. Our versatile visualization tool provides valuable insights into the characteristics of ViT models, offering substantial potential for practical applications. 

In Appendix~\ref{sec:additional_visualization} we offer additional visualization of ensembled SRT features across various network layers. The visualization indicates a noticeable trend: deeper layers reveal clearer high-level semantic boundaries,
while shallower layers highlight more local features than deeper ones.

\subsection{Semi-supervised Video Object Segmentation}\label{sec:davis}

\def\figd{figures/davis}
\def\fWidD{0.24\textwidth}
\begin{figure}[t]
\centering
{
\footnotesize
\hspace*{-2mm}
\begin{tabular}{c@{\hspace{0.05in}}c}
\includegraphics[width=\fWidD]{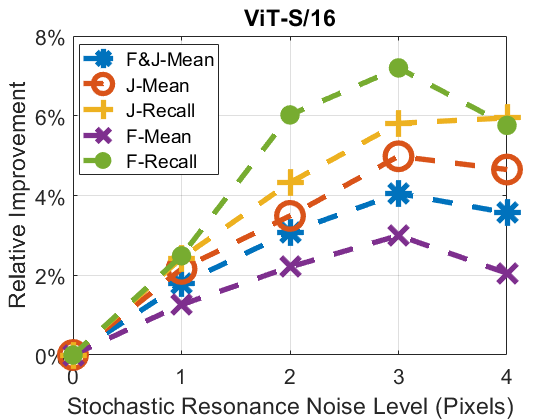}
&\includegraphics[width=\fWidD]{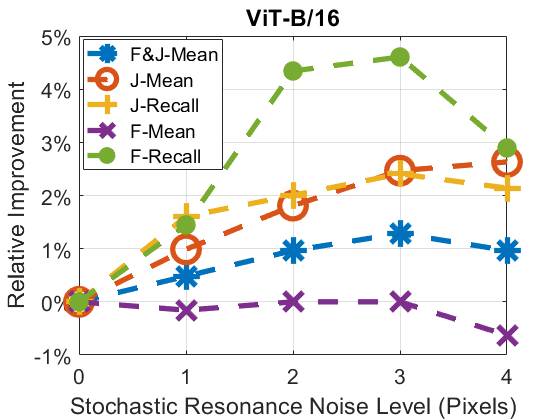} 
\end{tabular}
\caption{\sl\small {\bf Relative improvement on DAVIS-2017 dataset vs different noise levels.} There exists an inherent trade-off between perturbation level and performance gain. Smaller perturbation ranges result in weaker improvements from the baseline model due to lower input diversity, while larger perturbations are susceptible to greater information loss. 3 pixels is found to be the optimal perturbation level on both ViT-S/16 and Vit-B/16.}
}
\label{fig:noiselevel}
\end{figure}

We apply SRT to evaluate its performance using the DAVIS-2017 video instance segmentation benchmark \citep{Pont-Tuset_arXiv_2017}. We adhere to the experimental methodology established in \citep{jabri2020space}, which employs a "semi-supervised" video object segmentation approach on the original 480p resolution. Provided with the initial annotation of the objects of interest in the first frame, this method subsequently propagates the segmentation between consecutive frames. Notably, the method utilizes the last layer feature of the Vision Transformer (ViT) to guide this segmentation propagation process. Consequently, the quality of the ViT features directly impacts the final segmentation results. For optimal outcomes, these features must possess discriminative and semantically meaningful characteristics to effectively support this segmentation task.

In our study, we evaluate various Vision Transformer (ViT) models pre-trained using the DINO \citep{caron2021emerging} contrastive scheme. We adopt three different architectures, specifically ViT-S/16, ViT-B/16, and ViT-S/8, each varying in their spatial patch size (16x16 pixels and 8x8 pixels). Our results in Tab.~\ref{tab:davis} indicate that, on average, SRT enhances the original baseline models by a relative 2.4\% in terms of the F\&J score. The most significant improvement is observed with ViT-S/16, where we achieve 4.1\%. Importantly, these enhancements are achieved without any modifications to the model or pre-trained weights. However, we address a potential criticism of our approach, which could be seen as trivial test-time augmentation combined with feature-level ensemble. To counter this concern, we perform a heuristic by naively augmenting images by color jitter and performing feature-level ensemble (\ref{tab:davis}, Naive ensemble), and we find that this method is, in fact, detrimental to performance. We also reproduce the approach proposed by \cite{amir2021deep} that uses overlapping tokens at inference time, which negatively impacts the results. We investigate whether inference costs induced by SRT can potentially be mitigated via distillation. To this end, we attempt to learn the ensembled SRT representations using the following self-distillation objective:
\begin{equation}\label{eq:distill}
\min_w \sum_{x \in \mathcal{D}} ||\phi_w(x) - SRT(x, w_0)||,
\end{equation}
where $\phi$ and ($w_0$) $w$ are the ViT and its (original) parameters, and $x$ the image in the target dataset. Our preliminary results on DINO-ViT/16 improve from the baseline by $1.3\%$ after the self-distillation step. Note that Eq.~\eqref{eq:distill} is task agnostic and requires no label, thus effectively adapts pre-trained ViT features to any given target dataset.
We leave the investigation of this to future work.

Fig.~\ref{fig:noiselevel} illustrates the relative improvement across different perturbation levels of SRT applied to ViT-S/16 and ViT-B/16. While higher perturbation levels offer greater input diversity, they are also susceptible to information loss. We anticipate a trade-off between perturbation level and performance gain and empirically identify a perturbation level of 3 pixels as the optimal point for both.

\begin{table}[t]
\setlength{\tabcolsep}{2.7pt} 
  \centering
  \begin{tabular}{l|c|c|c|c|c}
    Method&F\&J&{\footnotesize J-mean}&{\footnotesize J-recall}&{\footnotesize F-mean}&{\footnotesize F-recall}\\    
    \hline
    DINO-ViT-S/16&0.617&0.602&0.740&0.634&0.764\\
    + SRT&\bf{0.642}&\bf{0.632}&\bf{0.783}&\bf{0.653}&\bf{0.819}\\
    \color{blue}{Distill by SRT} & \color{blue}{0.625} & \color{blue}{0.609} & \color{blue}{0.745} & \color{blue}{0.642} & \color{blue}{0.780} \\
    {\footnotesize + Overlapping tokens}&0.591&0.577&0.706&0.605&0.741\\
    + Naive ensemble&0.477&0.455&0.468&0.500&0.542\\
    \hline
    DINO-ViT-B/16&0.622&0.608&0.748&0.637&0.760\\
    + SRT&\bf{0.630}&\bf{0.623}&\bf{0.766}&\bf{0.637}&\bf{0.795}\\
    \hline
    DINO-ViT-S/8&0.706&0.675&0.815&0.737&0.846\\
    + SRT&\bf{0.720}&\bf{0.688}&\bf{0.827}&\bf{0.752}&\bf{0.868}\\
    \end{tabular}
  \caption{\sl\small{\bf Results on DAVIS-2017 video object segmentation.} Applying SRT improves over the baseline models uniformly over all metrics, as measured across 3 variants of ViTs trained using the DINO \citep{caron2021emerging} contrastive learning objective. SRT yields significant improvements even for ViT-S/8 trained with finer patch sizes (8x8). One may optionally fine-tune the original ViT model by {\color{blue} distilling} by SRT, which increases performance while inference time and cost remain one single forward pass.}
     \label{tab:davis}
\end{table}

\begin{table*}[t]
\setlength{\tabcolsep}{2.7pt} 
  \centering
  \begin{tabular}{l|l|l|c|c|c|c|c|c|c}
    Backbone&Head&Method&RMSE&RMSE\_log&AbsRel&SqRel&a1&a2&a3\\
    \hline\hline
    \multirow{6}{*}{DINOV2-ViT-B/14}&    \multirow{3}{*}{Linear}&Baseline&0.396&0.135&0.100&0.061&0.903&0.983&0.996\\
    &&+OE&0.376&0.121&0.093&0.059&0.918&0.984&0.997\\
&&+SRT&\bf{0.349}&\bf{0.108}&\bf{0.087}&\bf{0.052}&\bf{0.930}&\bf{0.990}&\bf{0.998}\\

\cline{2-10}
       & \multirow{3}{*}{DPT}&Baseline&0.323&0.109&0.074&0.044&0.941&0.987&0.996\\      &&+OE&0.314&0.101&\bf{0.073}&\bf{0.043}&0.944&0.988&\bf{0.997}\\
&&+SRT&\bf{0.305}&\bf{0.096}&\bf{0.073}&\bf{0.043}&\bf{0.945}&\bf{0.989}&\bf{0.997}\\

    \hline\hline
    \multirow{6}{*}{DINOV2-ViT-S/14}&    \multirow{3}{*}{Linear}&Baseline&0.471&0.162&0.125&\bf{0.084}&0.853&0.972&0.994\\
    &&+OE&0.486&0.153&0.126&0.095&0.858&0.974&0.994\\
&&+SRT&\bf{0.457}&\bf{0.140}&\bf{0.118}&0.085&\bf{0.876}&\bf{0.980}&\bf{0.996}\\

\cline{2-10}
&    \multirow{3}{*}{DPT}&Baseline&0.336&0.114&0.080&\bf{0.048}&0.933&0.986&0.996\\
&&+OE&0.347&0.114&0.080&0.053&0.932&0.985&0.996\\ 
&&+SRT&\bf{0.334}&\bf{0.104}&0.080&0.051&\bf{0.935}&\bf{0.988}&0.996\\

    \hline\hline
    \multirow{6}{*}{DINOV2-ViT-L/14}&    \multirow{3}{*}{Linear}&Baseline&0.373&0.127&0.093&0.054&0.916&0.985&0.996\\
&&+OE&0.401&0.131&0.097&0.062&0.908&0.982&0.996\\ 
&&+SRT&\textbf{0.365}&\textbf{0.113}&\textbf{0.090}&\textbf{0.053}&\textbf{0.924}&\textbf{0.989}&\textbf{0.998}\\

\cline{2-10}
&    \multirow{3}{*}{DPT}&Baseline&0.311&0.105&\bf{0.070}&0.042&0.946&0.988&\bf{0.997}\\
&&+OE&0.317&0.103&0.072&0.044&0.942&0.987&0.996\\ 
&&+SRT&\bf{0.297}&\bf{0.092}&\bf{0.070}&\bf{0.041}&\bf{0.947}&\bf{0.991}&\bf{0.997}\\

\end{tabular}
  \caption{\sl\small{\bf Results on NYU-V2 depth prediction.} Our method can be extended without modification to improve intermediate features to yield improved performance on the downstream depth prediction tasks. While ensembling of outputs (OE) can often be detrimental to performance, applying SRT on the features from pre-trained backbones (inputs to prediction heads) can improve performance over baselines by $4.7\%$ and $14.9\%$ on RMSE and RMSE\_log, using the linear prediction head and by $3.6\%$ and $11.0\%$ using the DPT head. }
     \label{tab:nyu}
\end{table*}

\subsection{Monocular Depth Prediction}
We extend the application of SRT to monocular depth estimation, a task that leverages ViT features from multiple ViT layers, in contrast to video object segmentation which primarily utilizes the last layer features. This choice of task highlights the versatility of SRT, showcasing its seamless compatibility with various ViT layers and architectures. Specifically, we evaluate three ViT architectures: ViT-S/14, ViT-B/14, and ViT-L/14, each equipped with two prediction heads (linear and DPT \citep{ranftl2021vision}). We adopt the experimental settings provided by DINOV2, which offers pre-trained backbones and corresponding prediction heads. Our assessment utilizes the NYU-V2 dataset \citep{NYUV2} under its original $640 \times 480$ resolution.

Tab.~\ref{tab:nyu} presents the results, demonstrating consistent improvements over baseline methods. The most significant enhancements are observed in the RMSE and RMSE\_log metrics, where we achieve relative improvements of 4.7\% and 14.9\% with linear heads, and 3.6\% and 11.0\% with DPT heads, respectively. Notably, these metrics are sensitive to outliers, highlighting the effectiveness of SRT in mitigating instability in ViT features and enhancing robustness.

For ablation, we compare our method with output-space ensemble (marked as "OE"), which employs the same perturbations as SRT, but aggregates the model output instead of intermediate features. We find no significant improvements, and in some cases, this method is even detrimental. This underscores the robustness of SRT's ensemble scheme that operates on the feature level instead of the output.

\subsection{Unsupervised Salient Region Segmentation}

We employ SRT in conjunction with TokenCut \citep{wang2022tokencut} for unsupervised salient region segmentation tasks. TokenCut is a graph-based approach that applies the Normalized Cut algorithm to partition ViT tokens into two distinct clusters, representing the salient foreground and the background respectively. The key challenge is to ensure that the features are not only discriminative across clusters but also consistent within clusters. We adopt three datasets: ECSSD \citep{shi2015hierarchical}, DUTS \citep{wang2017learning}, and DUT-OMRON \citep{yang2013saliency}, following the TokenCut. 

In Tab.~\ref{tab:saliency}, we report results both before and after post-processing (bilateral solver) to assess both the raw quality of ViT embeddings and final segmentation accuracy. Under both settings, SRT improves the original ViTs pre-trained by DINO, with an average increase in the maxF metric of 1.8\%. Notably, this improvement is constrained by the architecture of TokenCut, as it operates at the coarse segmentation level of ViT tokens. Directly applying TokenCut to the enhanced feature map is computationally impractical due to its $O(n^2)$ complexity in constructing a fully connected graph for graphcut. Given SRT's capability to provide fine-grained features, we anticipate future research on the effective leverage of SRT's high-resolution embeddings.

\begin{table*}[t]
\setlength{\tabcolsep}{2.7pt} 
  \centering
  \begin{tabular}{l|ccc|ccc|ccc}
    Datasets&\multicolumn{3}{c|}{ECSSD}&\multicolumn{3}{c|}{DUTS}&\multicolumn{3}{c|}{DUTS-OMRON}\\
      \hline\hline
      Feature Extractor&maxF&IoU&Acc.&maxF&IoU&Acc.&maxF&IoU&Acc.    \\
    \hline
DINO ViT-S/16&80.3&71.2&91.8&67.2&57.6&90.3&60.0&53.3&88.0\\
+SRT&\bf{82.4}&\bf{71.7}&\bf{92.1}&\bf{68.8}&\bf{58.5}&\bf{90.7}&\bf{61.0}&\bf{54.0}&\bf{88.2} \\
\hline
DINO ViT-S/16 w/ bilateral solver&87.4&\bf{77.2}&93.4&75.5&\bf{62.4}&91.4&69.7&61.8&89.7\\
+SRT&\bf{88.4}&77.0&\bf{93.6}&\bf{76.5}&\bf{62.4}&\bf{91.7}&\bf{70.6}&\bf{62.4}&\bf{89.9} \\
\hline
DINO ViT-B/16&80.3&71.0&91.5&66.4&56.7&89.5&56.7&50.5&85.4\\
+ SRT&\bf{81.8}&\bf{72.6}&\bf{92.2}&\bf{68.8}&\bf{58.3}&\bf{90.6}&\bf{58.0}&\bf{51.6}&\bf{86.1} \\
\hline
DINO ViT-B/16 w/ bilateral solver&86.8&76.6&93.0&74.1&60.9&90.6&65.6&58.4&87.1\\
+ SRT&\bf{88.2}&\bf{78.0}&\bf{93.7}&\bf{68.8}&\bf{58.3}&\bf{90.6}&\bf{67.2}&\bf{59.7}&\bf{87.8} \\    

\end{tabular}
  \caption{\sl\small{\bf Results on unsupervised salient region segmentation.} Despite architectural constraints, our method yields consistent improvement on all three datasets, with an average increase of 1.8\% in the maxF metric.
  }
     \label{tab:saliency}
\end{table*}

\subsection{Sanity Check: Image Retrieval and Unsupervised Object Detection}\label{sec:sanity}

Incorporating SRT into vision tasks involves updating ViT features based on fine-tuned high-resolution features. However, questions remain regarding whether the observed enhancements in dense prediction tasks are solely due to increased awareness of semantic boundaries in images and whether this method extends to non-dense prediction tasks. To address these concerns, we conducted a sanity check using image retrieval and unsupervised object detection tasks. 

For image retrieval, we applied a nearest-neighbor protocol following DINO, using the Oxford image retrieval datasets \citep{radenovic2018revisiting} and ViT-S/16 trained on ImageNet. Notably, our base model's pre-training poses a substantial domain gap to the target datasets. Note that, we do not naively average the class tokens from augmented images, but ensemble the features by SRT prior to the attention mechanism in the last layer. In this way, the final class token is computed from the ensemble SRT feature. Although image retrieval primarily requires distinctive image-level features (rather than pixel-level), aiming to match images to queries at a higher level, SRT exhibited effective adaptation, resulting in a notable 2.6\% relative improvement.

Regarding unsupervised object detection, we utilized TokenCut and the VOC07 dataset \citep{everingham2010pascal}. Unsupervised object detection focuses on region-level discriminative features, utilizing bounding boxes instead of segmentation masks for object shapes. Despite this, we observed a 1.0\% relative improvement in the detection rate, reaffirming that SRT does not compromise the information within the original ViT embeddings. These results serve as a critical validation of SRT's capacity to obtain fine-grained ViT features without distorting their original information.

\begin{table*}[t]
\setlength{\tabcolsep}{2.7pt} 
  \centering
  \begin{tabular}{l|l|c|c|c|c|c|c|c}
    Task&Metric&Baseline&d=1&d=2&d=3&d=4&d=5&d=6\\
      \hline
 \multirow{2}{*}{Image Retrieval}&mAP (Medium)&34.6&34.8&35.1&35.2&35.3&35.3&\bf{35.5}  \\
 &mAP (Hard)&13.0&13.1&13.2&13.1&13.2&\bf{13.2}&13.1\\
 \hline
 Object Discovery&Detection Rate&68.7&68.9&68.9&69.2&\bf{69.4}&69.3&69.2
\end{tabular}
  \caption{\sl\small{\bf Results on Image Retrieval and Object Discovery.} SRT generalizes to non-dense prediction tasks operating on higher-level region/image features to yield equal or better performance compared to the standard inference baseline. On the Oxford image retrieval task, SRT on the DINO-ViT-S/16 model yields up to $2.6\%$ relative improvement from the baseline model. On the unsupervised object detection task, SRT improves the detection rate by up to $1.0\%$. d: translation in pixels when ensembling with SRT. {\it d: perturbation level.}}
     \label{tab:results}
\end{table*}

\section{Related Work}\label{sec:related}
\textbf{Stochastic Resonance }
was proposed by \cite{benzi1981mechanism} and first applied in climate dynamics \citep{benzi1982stochastic} and later in  signal processing \citep{wellens2003stochastic, kosko2001robust, chen2007theory} and acoustics \citep{shu2016application, wang2014adaptive}. It is used to enhance a signal beyond the native resolution of the sensor by adding white noise. 
We use the same principle to adapt generic ViT image features for dense prediction downstream tasks. By randomly translating the images, (i.e. introducing noise in the spatial dimension), we can enhance ViT features to be smoother and better suited for dense prediction tasks. We leave extensions to other groups or semi-groups of transformations ({\em e.g.}, scale or domain size) to future work.


\textbf{Test-time data augmentation} involves aggregating model predictions from augmented test input to a final prediction. Applying such a technique increases the robustness of predictions \citep{prakash2018deflecting, song2017pixeldefend, cohen2019certified} and prediction accuracy \citep{krizhevsky2012imagenet, szegedy2015going,simonyan2014very, jin2018deep, matsunaga2017image} in a variety of tasks. It can also used to estimate the uncertainty of the model \citep{matsunaga2017image, smith2018understanding, ayhan2022test, wang2019aleatoric}. Different transformations are used to target different potential tasks: \cite{pang2019mixup} linearly combines the testing input and a randomly sampled clean image to generate classification prediction. \cite{isensee2018nnu} performs flipping and rotation to the test input image to generate 64 different inputs and finally aggregates the outputs to perform medical image segmentation. \cite{krizhevsky2012imagenet} crops the images into smaller patches and ensemble the results for classification.
Self-ensembling \citep{bousselham2021efficient} is also closely related to our work. \cite{bousselham2021efficient} leverages multi-scale features fed into multiple independent decoders to create an ensemble within a single model. \cite{liu2018towards} ensembles outputs from networks augmented with random noise layers to improve model robustness.  SRT aggregates information via adding spatial translations as noise and can be considered a general case of test-time augmentation, where ensembling is performed at the feature level at intermediate layers of a ViT, instead of the output level, which is novel.

\textbf{Knowledge distillation} aims to transfer the knowledge from stronger teacher models to weaker student models to improve their performance. \cite{hinton2015distilling} trains a student model to mimic the soft output distribution of the teacher model. \cite{romero2014fitnets} extends this idea to distill the intermediate features learned by the teacher models. We consider a form of self-distillation \citep{zhang2019your}, in which the student itself is used as the teacher to improve learned representations.

\textbf{Dense ViT feature extractor. } Our work is closely related to \citep{amir2021deep},which employs ViT for generating dense visual descriptors. To extract these fine-grained features, \citep{amir2021deep} reduce the stride allowing for overlapping tokens and performing a single forward pass with ViT. In SRT, instead of a single pass, we conduct multiple passes using perturbed inputs. This modification reduces the computational complexity from quadratic to linear.

\textbf{Properties discovered by SRT.} Additionally, our findings underscore the segmentation capabilities of ViT embeddings, aligning with recent claims in the field \citep{caron2021emerging,yu2023emergence}. Enhanced features exhibit sharp, fine-grained semantically relevant boundaries. Furthermore, our method leverages the convexity properties \citep{park2022vision} of ViT embeddings, enabling convex combinations (average pooling as a special case) during inference, resulting in improvements across various tasks. 

\section{Discussion}\label{sec:conclusion}
\noindent\textbf{Ensemble vs super-resolution. } Although both increase the spatial resolution, SRT achieves it by ensemble, which differs from super-resolution: Given a signal $x$ that is sub-sampled to $\tilde x$, super-resolution aims to retrieve an approximation $\hat x$ of $x$ given $\tilde x$. Since information is lost in the sampling, the reconstruction depends crucially on the choice of prior. In this sense, super-resolution is a form of hallucination: Attribute details to $\hat x$ that are not in $\tilde x$, in the hope that they will somehow match those in $x$. This requires strong faith in prior knowledge about $x$, $P(x)$. 

Given the same signal $x$, one could instead generate multiple samples $\tilde x_i$, each with a different kernel, and then reconstruct a single estimate from the samples $\hat x = F(\tilde x_1, \dots, \tilde x_N)$. Now, the estimator $F$ has more information available about $x$ than in super-resolution: The sigma-algebra spanned by the random variables $\tilde x_i$ is a superset of the (trivial) sigma algebra spanned by the single sample $\tilde x$ in super-resolution.

In other words, SRT which aggregates different samples from a process contains more information than any single sample about the process. We provide further ablation studies and discussions on comparing with enhancing the features by spatial interpolation and smoothing in Appendix~\ref{sec:interp}.

\noindent\textbf{Limitations. } SRT has several limitations. The basic embodiment increases inference cost and latency, as each perturbed image necessitates a ViT forward pass. To address this, one viable approach is knowledge distillation, which involves fine-tuning the network to mimic the feature-level output of SRT.
We illustrate this process using the DAVIS-2017 training dataset with DINO-ViT-S/16, achieving improved results (F\&J-score $0.617 \Rightarrow 0.625$) without the use of labels or operations on the validation set. This establishes a label-free, task-free transductive fine-tuning scheme that adapts pre-trained ViT features to new target datasets. Future directions may involve refining the distillation process on different layers and exploring the integration of Stochastic Resonance directly into ViT architectures.

\noindent\textbf{Conclusions. } SRT offers a versatile feature-level ensemble method that applies to any layer within any architecture that utilizes ViT as a feature extractor, eliminating the need for modifications to the forward pass, in contrast to most test-time augmentation and ensemble methods that operate at the output level that require task-specific designs. Compared with increasing token numbers, SRT avoids the quadratic complexity related to the number of ViT tokens, and is amenable to parallelization through batching, ensuring computational efficiency. Furthermore, the method allows ensembling without memory-intensive resizing of all embeddings to full resolution, which can be executed recursively, as described in Sect.~\ref{sec:efficient}. Practical implementations demonstrate efficient execution on even laptop GPUs. 

It is worth noting that stochastic resonance is not limited to ViT architectures, as demonstrated in Appendix~\ref{sec:resnet}, where we apply the mechanism to ResNet on the image classification task, and on average reduce error by a relative 5.87\%. Stochastic resonance also applies to other forms of quantization, such as sale or domain size. However, our emphasis in this paper is on ViTs that mostly use non-overlapping tokens, making them particularly suited to our approach. 

\section{Impact Statement}
This paper presents work whose goal is to advance the field of Machine Learning. There are many potential societal consequences of our work, none of which we feel must be specifically highlighted here.

\comment{

\subsection{Distillation}
\begin{table}[h]
\setlength{\tabcolsep}{2.7pt} 
  \centering
  \begin{tabular}{l|c|c|c|c|c}
    Method&F\&J-Mean&J-Mean&J-Recall&F-Mean&F-Recall\\    
    \hline
    DINO-ViT-S/16&0.617&0.602&0.740&0.634&0.764\\
    + SRT&\bf{0.642}&\bf{0.632}&\bf{0.783}&\bf{0.653}&\bf{0.819}\\
    + Distill & 0.625 & 0.609 & 0.745 & 0.642 & 0.780 \\
    \end{tabular}
  \caption{\sl\small{\bf Distillation Results on DAVIS-2017 video object segmentation.} We show that the SRT representations can be learned by the base model via standard knowledge distillation objectives. This method of self-distillation can be used to improve a model at inference time in an unsupervised manner, similar to the objectives of ``test-time training" \citep{sun2020test}.}
     \label{tab:distill}
\end{table}

In this section, we investigate whether the coarse granularity of the representations learned are a fundamental limitation of the network architecture. If not, inference cost induced by SRT can potentially be mitigated via distillation. Towards this goal, we attempt to learn the ensembled SRT representations using the following self-distillation objective:
\ty{notation}
\begin{equation}
\min_w \sum_{x \in \mathcal{D}} ||\phi_w(x) - SRT(x, w_0)||
\end{equation}
Our preliminary results in Tab.~\ref{tab:distill} suggest that the coarseness of the representations learned emerges from the pre-training objective, rather than being a consequence of the patch-based architecture, improving results from the baseline by $X\%$ after the self-distillation step. This suggests that our method can be applied to perform ``test-time training" \citep{sun2020test} that evolves the model after each inference request. We leave the investigation of this to future work.


\subsection{Limitations and Future Work}
    
}

\bibliography{iclr2024_conference}
\bibliographystyle{icml2024/icml2024}
\clearpage

\appendix
\onecolumn
\section{Results on CNNs and Classification}\label{sec:resnet}

\begin{table}[h]
\setlength{\tabcolsep}{2.7pt} 
  \centering
  \begin{tabular}{l|c|c|c}
    Architecture&ResNet20&ResNet32&ResNet56\\
      \hline\hline
      Accuracy&91.95&92.68&93.50\\
      Accuracy w/ SRT&\bf{92.41}&\bf{93.14}&\bf{93.87}
      \\
      \hline
      Relative error reduced&5.6\%&6.3\%&5.7\%
\end{tabular}
  \caption{\sl\small{\bf Results on Cifar-10 classification with ResNet.} Stochastic resonance consistently improves classification accuracy by an average of 5.87\% and as much as 6.3\% on ResNet32, without additional training. 
  }
     \label{tab:resnet}
\end{table}

As mentioned in the paper, Stochastic Resonance is not confined to Vision Transformer architectures. We specifically opted for ViT and zero-shot methods to effectively showcase its benefits. In this context, we present additional results involving Convolutional Neural Networks (CNNs) and supervised image classification. We test ResNet~\cite{he2016deep} on the CIFAR dataset, and report the results in Tab.~\ref{tab:resnet}. Through the application of stochastic resonance, we employ ensembling at the final layer before the prediction head. Across ResNet20, ResNet32, and ResNet56, we consistently observe improvements, with the prediction error reduced at inference time by an average of average of 5.87\% and as much as 6.3\%, all without the need for additional training.

\section{Results on Semantic Segmentation}\label{sec:semantic}
\begin{table}[h]
  \centering
  \begin{tabular}{l|l|c|c|c|c}
    Method&head&baseline&d=1&d=2&d=3\\
      \hline
 DINOV2 ViT-S/14&linear& 44.24 & 44.44 & 44.57 & \bf{44.64}  \\
 \hline
 DINOV2 ViT-B/14&linear& 47.28 & 47.63 & 47.85 & \bf{47.98} \\
 \hline
 DINOV2 ViT-L/14&linear& 47.79 & 48.18 & 48.44 & \bf{48.62}
\end{tabular}
  \caption{\sl\small{\bf Results on Semantic Segmentation on ADE20K in mIOU} Experiments run with evaluation pipeline from InternImage \citep{Wang_2023_CVPR} and DINOV2 \cite{oquab2023dinov2}. d denotes the translation in pixels, ranging from -d to d with respect to a coordinate location across horizontal and vertical directions, when ensembling with SRT. As the size of the ensemble grows, the segmentation mIOU increases.}
     \label{tab:semantic}
\end{table}

We show results on combining semantic segmentation with SRT. We employ the protocol from DINOV2~\cite{oquab2023dinov2} and ADE20K dataset~\cite{zhou2017scene}. The results are presented in Tab.~\ref{tab:semantic}. SRT consistently improves mIoU on all three pre-trained ViTs, by as much as 1.7\% in relative improvement. In comparison, results on depth prediction show a more significant improvement (14.9\%). We conjecture that depth prediction benefits more from SRT as it is a geometry task and SRT leverages a geometric augmentation. Nevertheless, the gain in semantic segmentation is obtained without any further fine-tuning.

\section{Discussion: Comparison with Feature Interpolation}\label{sec:interp}

One might speculate that the performance improvement of SRT arises from a fine-grained feature map, which may be advantageous for dense prediction tasks. We conduct a sanity check in Sect.~\ref{sec:sanity} in the main paper, and here we further compare SRT with resizing the feature field through interpolation. We adopt the depth prediction task using NYU-V2, consistent with the main paper, and the results are presented in Tab.~\ref{tab:interpolation}. Somewhat surprisingly, the simple approach of interpolating the feature map leads to a performance decrease. Two possible explanations are considered. First, the architecture and training loss lack an explicit constraint on the smoothness of features, making spatial interpolation problematic. Second, while interpolation increases feature resolution, it does not introduce new information, whereas ensembling effectively samples the input signal (augmented image) multiple times, which contains more information than any single sample.

\begin{figure}
    \centering
    \includegraphics[width=\textwidth]{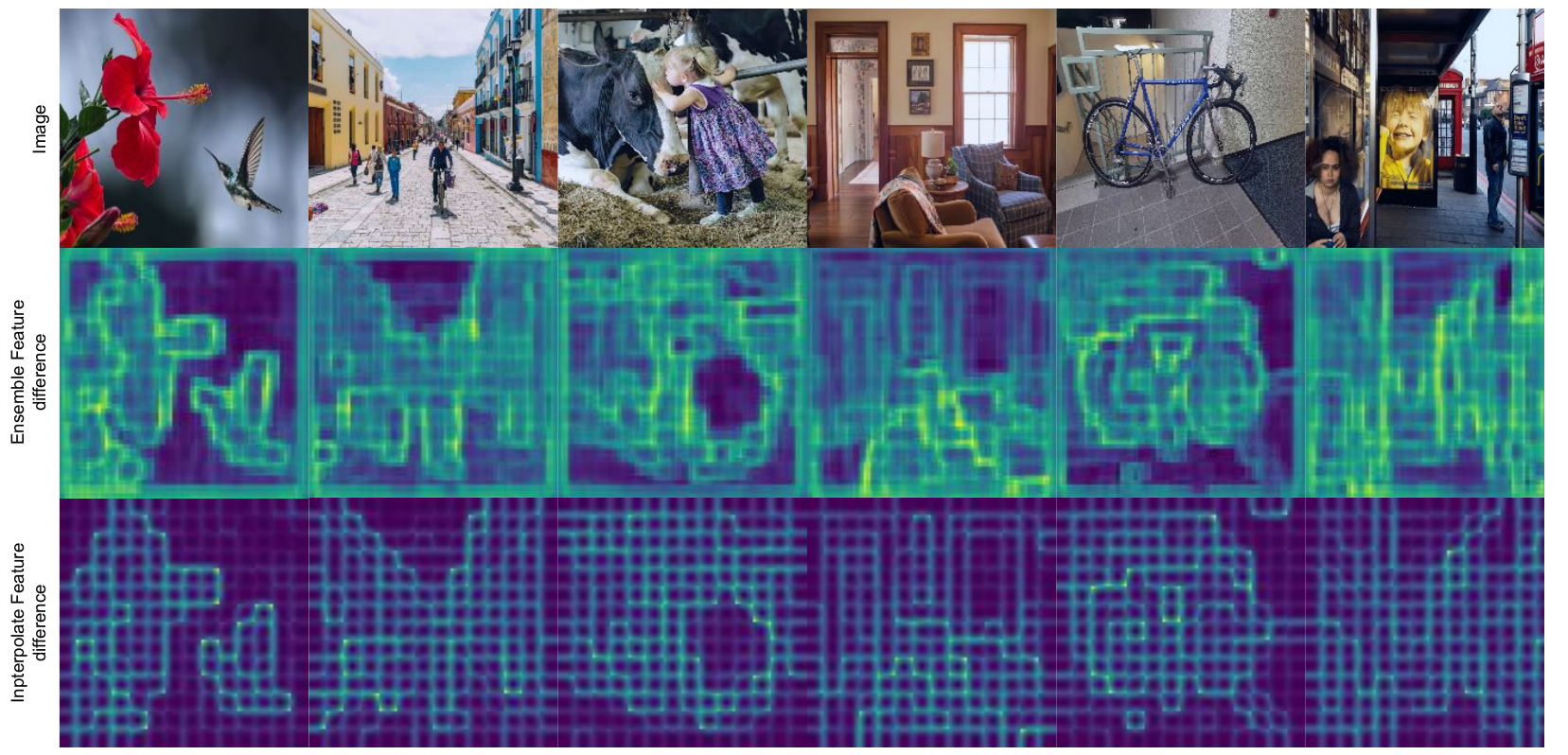}
    \caption{\sl\small{\bf Noise distribution in the features by SRT.} Considering the ensembled features represent a "denoised" signal, we visualize noise distribution, which aligns to semantic boundaries (2nd row) where image patches do not align with object shape due to quantization. For reference, we also show the difference between resized features (by interpolation) to the original feature, which shows a less meaningful grid pattern.}
    \label{fig:noise-distribution}
\end{figure}

Considering the ensembled features represent a "denoised" signal, we can measure the noise distribution in the features by the L2 difference between the ensembled features and the feature from a single forward pass, aggregated as a histogram. Fig.~\ref{fig:noise-distribution} visualizes this noise distribution. It is easy to notice that, the noise is mostly aligned to the semantic boundaries where image patches do not align with object shape due to quantization. In comparison, we also visualize the difference between the resized features (by interpolation) and the single-forward-pass feature. The resulting noise map is a less meaningful grid-like structure due to the image quantization and interpolation artifacts. Note that the center of each patch of the difference maps of the interpolated features is ``dark'' meaning there is no information introduced. On the contrary, the difference maps of SRT is ``bright'', which comes from the noise introduced by Stochastic Resonance to enhance the signal.

\begin{table}[t]
\setlength{\tabcolsep}{2.7pt} 
  \centering
  \begin{tabular}{l|l|l|c|c|c|c|c|c|c}
    Backbone&Head&Method&RMSE&RMSE\_log&AbsRel&SqRel&a1&a2&a3\\
    \hline\hline
    \multirow{3}{*}{DINOV2-ViT-S/14} & \multirow{3}{*}{DPT} &  Baseline & 0.336 & 0.114 & 0.080 & 0.048 & 0.933 & 0.986 & 0.996 \\
    && Bilinear & 0.573 & 0.178 & 0.146 & 0.125 & 0.8	& 0.964 & 0.995 \\
    & & Bicubic & 0.572 & 0.124 & 0.146 & 0.178 & 0.801 & 0.964 & 0.995 \\
    \cline{1-10}
    \multirow{3}{*}{DINOV2-ViT-B/14} & \multirow{3}{*}{DPT} & 
    Baseline & 0.323 & 0.109 & 0.074 & 0.044 & 0.941 & 0.987 & 0.996 \\
    &&Bilinear & 0.568 & 0.185 & 0.146 & 0.120 & 0.792 & 0.96 & 0.992 \\
    & & Bicubic &  0.579 & 0.188 & 0.149 & 0.124 & 0.787 & 0.959 & 0.991\\
    \cline{1-10}
     \multirow{3}{*}{DINOV2-ViT-L/14} & \multirow{3}{*}{DPT} & 
     Baseline & 0.311 & 0.105 & 0.070 & 0.042 & 0.946 & 0.988 & 0.997 \\
     & & Bilinear & 0.732 & 0.246 & 0.183 & 0.195 & 0.695 & 0.91 & 0.973 \\
     & & Bicubic & 0.720 & 0.241 & 0.181 & 0.188 & 0.701 & 0.916 & 0.975 \\
\cline{1-10}
\end{tabular}
  \caption{\sl\small{\bf Results on NYU-V2 depth prediction using interpolated features} For the interpolation method, we bilinearly or bicubically interpolate the DINOV2 feature up to the image dimension and perform an average pooling to return the feature to the original dimension. The results are much worse than the baseline method, which uses the original DINOV2 features without ensembling. }
     \label{tab:interpolation}
\end{table}

\begin{figure}
    \centering
    \includegraphics[width=\textwidth]{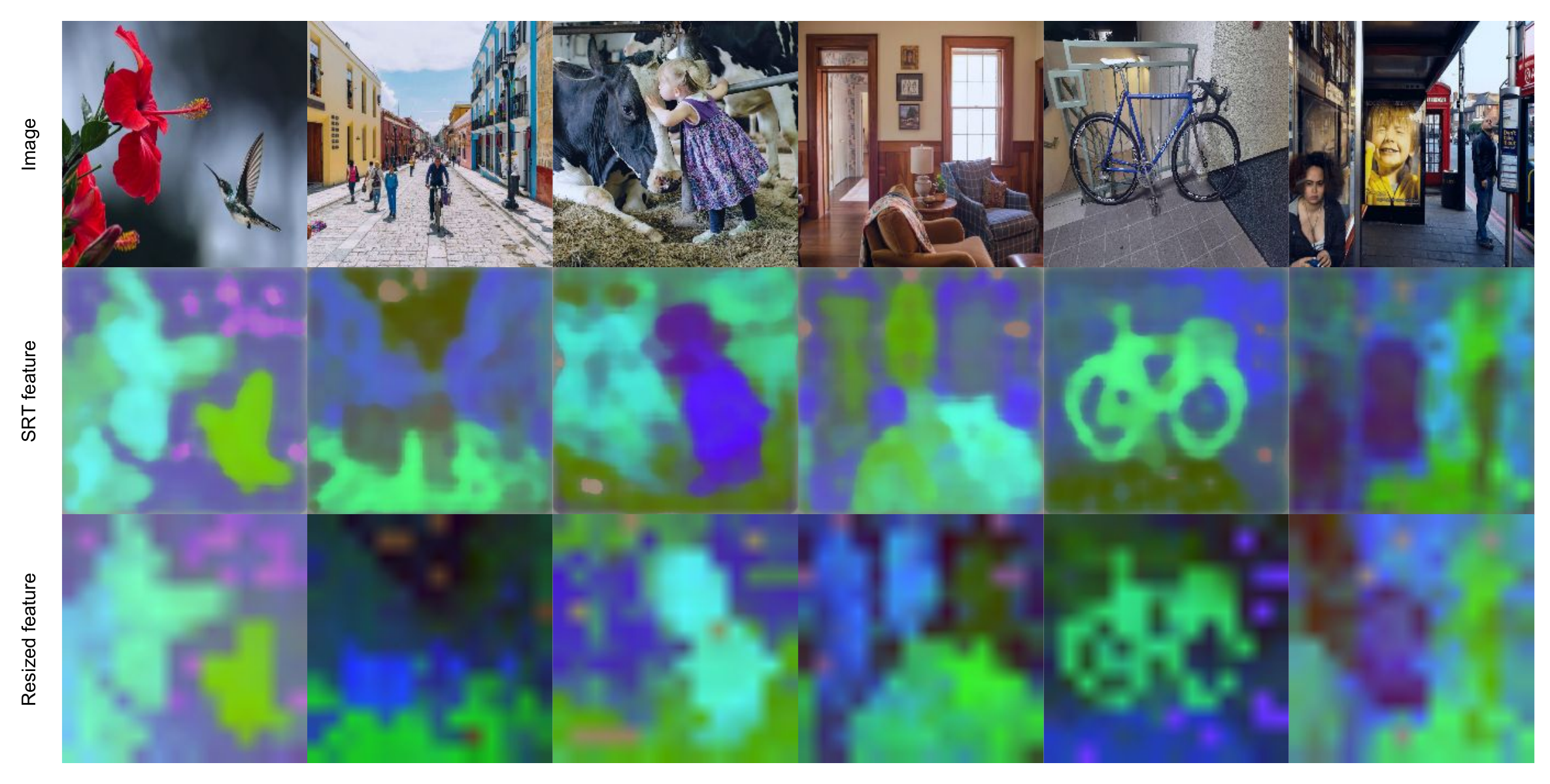}
    \caption{\sl\small{\bf Comparing SRT features and resized single-forward-pass features.} SRT features respect the semantic boundaries better than resized features (The edge of the bird and pedals in column one and wheels of the bicycle in column five). Resized features contain quantization artifacts where edges are vertical and horizontal lines corners are right angle corners. Our feature can represent much more detailed object contours. }
    \label{fig:srt-resize-comp}
\end{figure}

\section{Formal relation between SRT-induced representations and explicit models of image formation}\label{sec:formalization_appendix}

Stochastic Resonance is a phenomenon whereby {\em ``increases in levels of unpredictable fluctuations -- {\em e.g.}, random noise -- cause an increase in a metric of the quality of signal transmission or detection performance, rather than a decrease ''} \citep{mcdonnell2009stochastic} The phenomenon was first analyzed in reference to bi-stable systems of stochastic differential equations, where the variance of the observations decreased after injecting noise into the system \citep{benzi1981mechanism}. Subsequently, the idea was applied to quantized systems, where each level $x^i \in X = \{x^1, \dots x^N\}$ of a quantization operator $f: {\mathbb R} \rightarrow X \subset {\mathbb R}$ replaced the role of each stable attractor in the original formulation, and stochastic resonance resorted to averaging noisy qantized versions of the original signal $x$, 
\begin{equation}
    \hat x = \int f(x+n)dP(n)
\end{equation}
resulting in smaller noise variance ${\mathbb E}(|\hat x - x|^2)$  (or higher signal-to-noise ratio) than the original measured signal without added noise, 
${\mathbb E}(|x - f(x) |^2)$ where $\tilde x = f(x)$, all assuming sufficiently small perturbations $n \sim dP$ from a chosen distribution $P$ (a design choice). This method was successfully used in cochlear signal processing where  $f$ is a coarse quantizer implemented using low-power electronics. 

Now consider the more general case where $x$ is not just a real number but an image (irradiance function) defined on a compact planar domain,  $x: D \subset {\mathbb R}^2 \rightarrow {\mathbb R}; (u,v) \mapsto x(u,v)$; $f$ is a complex operator that, in addition to quantizing the planar domain $D$ with a function $\pi: D \rightarrow \Lambda$, where $\Lambda$ is a discrete lattice, also maps each cell in the lattice in a $K$-dimensional vector $\phi(\{x(u,v), (u,v) \in D_i\}) \in {\mathbb R}^K$, where $D_i$ is a cell in the lattice, so that $f: D\rightarrow {\mathbb R}^K; x \mapsto f(x)$ with
\begin{equation}
f(x)= \phi(\pi(x))
\label{eq:pi_projection}
\end{equation}

where $\pi$ denotes the restriction to the lattice. Now, instead of considering an additive perturbation that acts on the range space of $x$, $x \mapsto x+ n$, we consider a more general perturbation operator $T$ that can act on either the domain $D$ or the range $x$, which we indicate with $Tx$. This can be a trivial additive perturbation, $T(n)x = x + n$ for some scalar $n$, or it can be a planar translation $T(n)x(u,v) = x(n+ n_u, v + n_v)$, where $n = (n_u, n_v)$ is a translational offset, or it can be an affine, projective, diffeomorphic or homeomorphic deformation of both the domain and the range of $x$. In this paper, we restrict ourselves to planar translation but the concepts extend to any invertible operator $T$. We write this formally as
\begin{equation}
    \hat x = \int f(Tx(u,v))dP(T(u,v))
    \nonumber
\end{equation}
which boils down to spatial averaging if we choose $dP$ to be constant ($P$ uniform). One can also consider scale averaging, which gives rise to so-called domain-size pooling \citep{dong2015domain}. In a discrete setting, the perturbation can be quantized and averaged
\begin{equation}
    \hat x = \sum_i \phi(\pi(x_i))
    \nonumber
\end{equation}
which can done sequentially as a moving average. Our model includes a slightly more sophisticated (forward-backward) projection operator $\pi$, that allows us to express the averaging in terms of the measured signal rather than the (unknown) original analog signal, since the latter is unknown.

Now, in addition to reducing the variance of the reconstruction error, what are the specific artifacts that arise in the presence of spatial quantization that we wish to recuperate by the use of perturbation-averaging? 

Consider an elementary image-formation model where a physical scene is composed of piecewise smooth multiply-connected surfaces, $S \subset {\mathbb R}^3$, each supporting a piecewise smooth radiance function (``texture'' or albedo) $\rho: S \rightarrow {\mathbb R}$, imaged through a pinhole (central) projection $\pi: {\mathbb R}^3 \rightarrow {\mathbb R}^2$, where the projection operator also includes spatial quantization into the lattice. Now, we have: 
\[
\tilde x(\tilde u, \tilde v) = \int_{\pi^{-1}(\tilde u, \tilde v)} \int_L \beta_{p(\tilde u, \tilde v)}(\tilde u - u, \tilde v-v)dE(u,v); \quad [\tilde u, \tilde v] = \pi(p), \ p\in S
\]
where $\beta$ is a bi-directional reflectance distribution function (whose integral over a unit solid angle around $(u,v)$ at each point $p$ yields the diffuse albedo $\rho$) and the integral over the light source $L$ extends to the pre-image under $\pi$ of the quantization cell around $(\tilde u, \tilde v)$ (this is the intersection of the cone subtending a patch centered at $(\tilde u, \tilde v)$ with reflective surfaces in the scene). Notice the inverse projection in the domain of integration and the forward projection in the selection of the projection point, corresponding to a cell in the lattice. 

One should also notice that the representation $\phi(x)$ computed at location $(\tilde u, \tilde v)$ does not just use the pixel at that location, nor does it simply average pixels in its neighborhood, but rather aggregates information from all pixels in the patch. Nonetheless, the information about the {\em scene} that is being aggregated changes with the distance of the scene, and translating the patch ({\em e.g.,} computing the representation at an adjacent patch, even if overlapping) mixes the contribution of different connected components of the underlying surface $S$, and corresponding segments of the albedo $\rho$. Therefore, even if vectorial, the representation $\phi$ is subject to quantization artifacts {\em when viewed as a representation of the scene, rather than of the given patch}, which causes artifacts such as the loss of details at occluding boundaries and albedo boundaries. By aggregating multiple samples at different values of the transformation $T$, we can recouperate some of those details, up to the quantization limits of the sampling of the transformation (as opposed to the quantization limits of the patch-based tokenization).

Generally, the quantized signal $\tilde x$ is piecewise constant, but the discontinuities correspond to the lattice cell boundaries, and have nothing to do with either the geometric discontinuities due to the piecewise smooth nature of $S$, or the photometric discontinuities due to albedo boundaries in $\beta$ or the corresponding $\rho$. As a result, object boundaries (which correspond either to occlusion/geometric boundaries, or material/albedo boundaries) are not visible in the quantized signal and generally can appear and disappear even at constant quantization levels simply by moving farther and closer to the camera due to the varying size of the intersection of the cone $\pi^{-1}(D_i) \cap S$. This gives rise to genetic effects of the kind familiar in scale space theory \citep{lindeberg2013scale}. 

While it would be ideal to be able to prove analytically that discontinuities due to material and illumination boundaries that are washed out by spatial quantization are recovered by stochastic resonance, even the simplistic image formation model above is way beyond the complexity that is amenable for direct analysis. For this reason, in the paper we resort to empirical tests, either qualitative by direct visualization, or qualitative by using the averaged feature $\hat x$ instead of the original feature $\tilde x$, in downstream inference tasks.

\section{Additional Visualization}\label{sec:additional_visualization}
We offer additional visualization of ensembled SRT features across various network layers in Fig.~\ref{fig:vit_layers}, using CLIP and DINO for illustration. Our visualization indicates a noticeable trend: deeper layers reveal clearer high-level semantic boundaries, while shallower layers highlight more local features compared to high-level ones.

\def\fWidD{0.125\textwidth}
\def\figd{figures/additional_vis}
\begin{figure*}[h]

\centering
{\footnotesize 
\begin{tabular}{c@{\hspace{0.05in}}c@{\hspace{0.05in}}c@{\hspace{0.05in}}c@{\hspace{0.05in}}c@{\hspace{0.05in}}c@{\hspace{0.05in}}c}
& & CLIP & & & DINO & \\
Image&layer 1&layer5&layer 9&layer 1&layer5&layer 9
\\
\includegraphics[width=\fWidD]{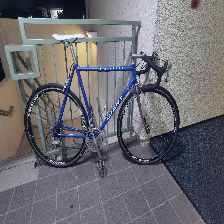}&\includegraphics[width=\fWidD]{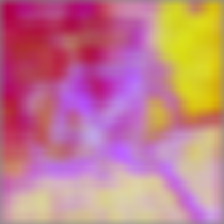}&\includegraphics[width=\fWidD]{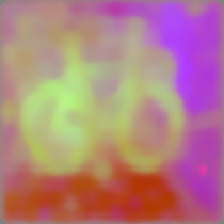}&\includegraphics[width=\fWidD]{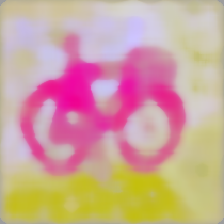}&\includegraphics[width=\fWidD]{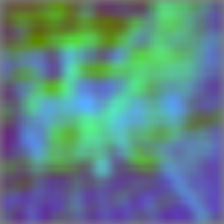}&\includegraphics[width=\fWidD]{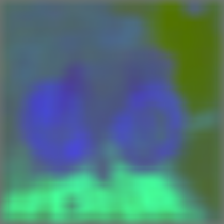}&\includegraphics[width=\fWidD]{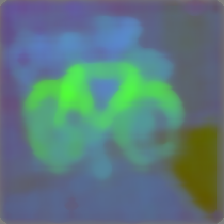}\\
\includegraphics[width=\fWidD]{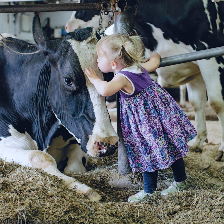}&\includegraphics[width=\fWidD]{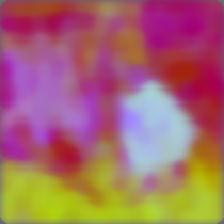}&\includegraphics[width=\fWidD]{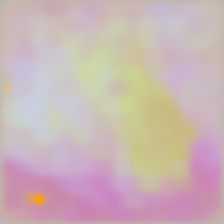}&\includegraphics[width=\fWidD]{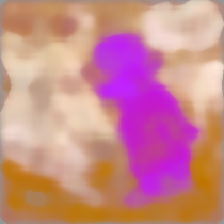}&\includegraphics[width=\fWidD]{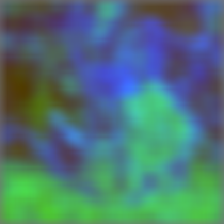}&\includegraphics[width=\fWidD]{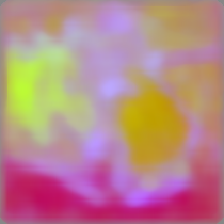}&\includegraphics[width=\fWidD]{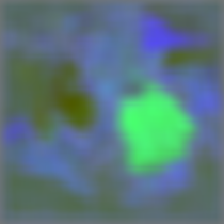}\\
\includegraphics[width=\fWidD]{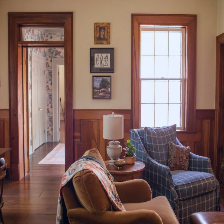}&\includegraphics[width=\fWidD]{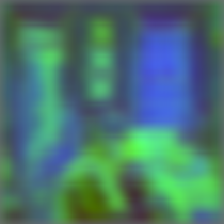}&\includegraphics[width=\fWidD]{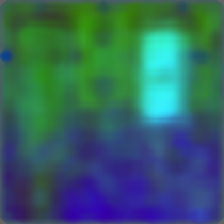}&\includegraphics[width=\fWidD]{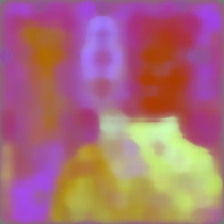}&\includegraphics[width=\fWidD]{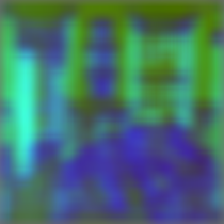}&\includegraphics[width=\fWidD]{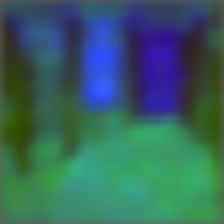}&\includegraphics[width=\fWidD]{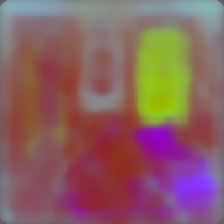}\\
\end{tabular}
}

\caption{\sl\small {\bf Visualization of ensembled SRT features in different ViT layers.} Architecture: ViT-S/16.}
\label{fig:vit_layers}
\end{figure*}

\comment{
    
\begin{algorithm}[t]
	\begin{algorithmic}[1]
		\REQUIRE{ViT feature extractor $f_\theta$, Image $I \in \mathbb{R}^{3\times H \times W}$, Level of pertubation $d\in \mathbb{N}_+$}
            \STATE $b \gets \{0\}^{N \times H \times W} $, $N$ the dimension of ViT features.
            \FOR{$x \in \{-d, \dots, 0, \dots, d\}$} 
                \FOR{$y \in \{-d, \dots, 0, \dots, d\}$} 
                    \STATE Translate $I$ by $x$ and $y$ pixels in the horizontal and vertical direction respectively to get $I'$
                    \STATE Compute ViT features for $I'$: $f_\theta(I')$
                    \STATE Upsample $f_\theta(I')$ to $f^{'}_\theta(I')\in \mathbb{R}^{N \times H\times W}$
                    \STATE Translate $f^{'}_\theta(I')$ by $-x$ and $-y$ in the horizontal and vertical direction respectively to obtain $f^{''}_\theta(I')\in \mathbb{R}^{N \times H\times W}$
                    \STATE Aggregate $f^{''}_\theta(I')$ into $b$
                \ENDFOR
            \ENDFOR
            \RETURN $\frac{b}{n_b}$, where $n_b$ is the number of features aggregated into $b$ 
	\end{algorithmic}
	\caption{\textsc{Algorithm to obtain SRT features.}}
	\label{alg:main-algorithm} 
 
\end{algorithm}

}

\end{document}